\documentclass[letterpaper, 10 pt, conference]{ieeeconf}

\IEEEoverridecommandlockouts                          
\usepackage{amsmath} % assumes amsmath package installed
\usepackage{amssymb}  % assumes amsmath package installed
\usepackage{algorithm}
\usepackage{algpseudocode}
\usepackage{graphicx}
\usepackage{physics,bm,bbm}

\usepackage{multirow}
\usepackage{cuted}
\usepackage{cite}

\usepackage{booktabs}
\usepackage{colortbl}
\usepackage{multicol}
\usepackage{xcolor}
\usepackage{makecell}
\usepackage{subcaption}
\usepackage{enumitem}
\newcommand{\hide}[1]{}

\definecolor{es-blue}{rgb}{0,0.4,0.8}

\definecolor{lightgray}{gray}{0.9}

\definecolor{cvprblue}{rgb}{0.21,0.49,0.74}
\usepackage[pagebackref,breaklinks,colorlinks=true,citecolor=cvprblue, urlcolor=blue]{hyperref}

\title{\LARGE \bf
Human-Robot Copilot for Data-Efficient Imitation Learning
}

\author{
    Rui Yan\textsuperscript{*} \quad
    Zaitian Gongye\textsuperscript{*} \quad
    Lars Paulsen \quad
    Xuxin Cheng \quad
    Xiaolong Wang \\
    UC San Diego
    \thanks{* Equal contributions.}
}

\begin{document}
\twocolumn[{%
\renewcommand\twocolumn[1][]{#1}%
\maketitle

\begin{center}
    \vspace{-0.1in}
    \centering
    \captionsetup{type=figure}
    \includegraphics[width=1\linewidth]{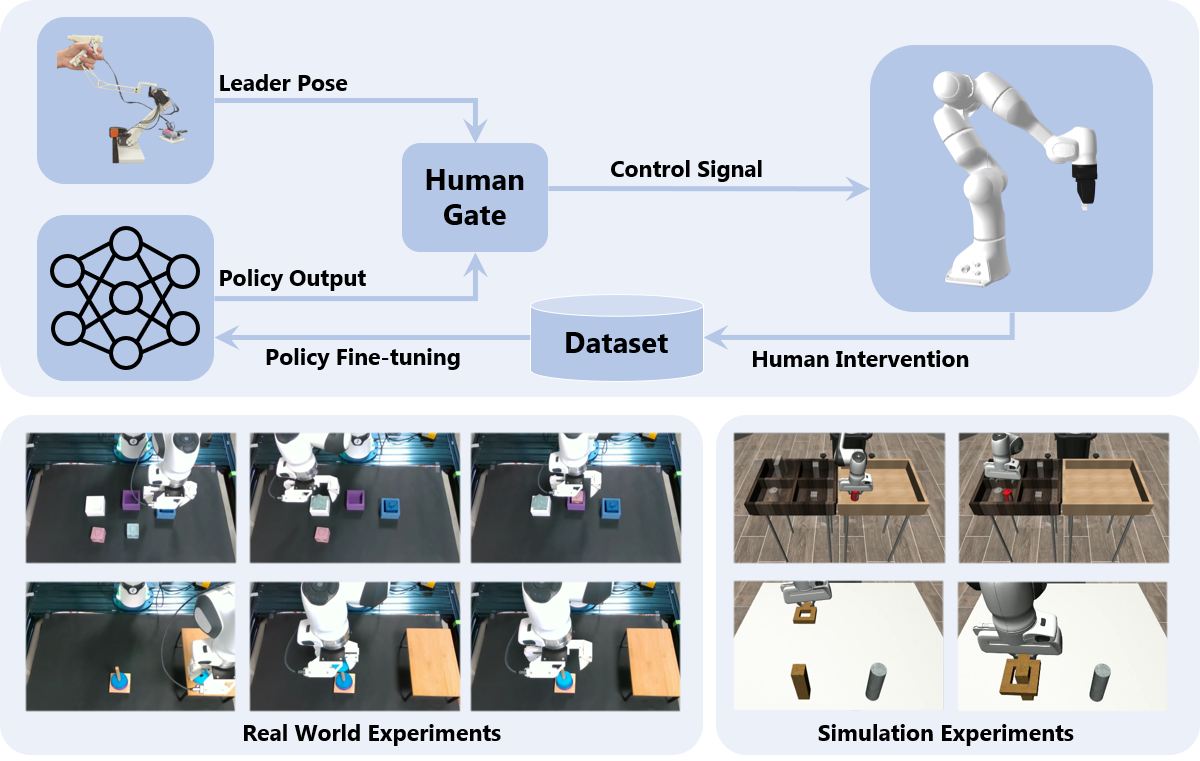}
    \caption{Overview of the Human-Robot Copilot framework. The human teleoperator determines when to intervene in policy execution and collect augmentation data for policy fine-tuning. The bottom right illustrates simulation experiments in robosuite, and the bottom left shows real-world experiments: cube sorting in a highly randomized environment and tower of Hanoi insertion requiring high-precision actions.}
    \label{fig:tease}
    \vspace{0.1in}
\end{center}
}]

% {\blfootnote{{\authorrefmark{1} Equal contribution. }}}

% \blfootnote{
% \\
% \textsuperscript{$1$} University of California San Diego \\
% \textsuperscript{$2$} Tsinghua University \\
% \textsuperscript{$\dag$} Work done during internship at UC San Diego.\\
% \textsuperscript{*} Equal contributions. \\
% }

\thispagestyle{empty}
\pagestyle{empty}

%%%%%%%%%%%%%%%%%%%%%%%%%%%%%%%%%%%%%%%%%%%%%%%%%%%%%%%%%%%%%%%%%%%%%%%%%%%%%%%%
\begin{abstract}

Collecting human demonstrations via teleoperation is a common approach for teaching robots task-specific skills. However, when only a limited number of demonstrations are available, policies are prone to entering out-of-distribution (OOD) states due to compounding errors or environmental stochasticity. Existing interactive imitation learning or human-in-the-loop methods try to address this issue by following the Human-Gated DAgger (HG-DAgger) paradigm, an approach that augments demonstrations through selective human intervention during policy execution. Nevertheless, these approaches struggle to balance dexterity and generality: they either provide fine-grained corrections but are limited to specific kinematic structures, or achieve generality at the cost of precise control. To overcome this limitation, we propose the Human-Robot Copilot framework that can leverage a scaling factor for dexterous teleoperation while maintaining compatibility with a wide range of industrial and research manipulators. Experimental results demonstrate that our framework achieves higher performance with the same number of demonstration trajectories. Moreover, since corrective interventions are required only intermittently, the overall data collection process is more efficient and less time-consuming.

\end{abstract}
%%%%%%%%%%%%%%%%%%%%%%%%%%%%%%%%%%%%%%%%%%%%%%%%%%%%%%%%%%%%%%%%%%%%%%%%%%%%%%%%
\section{Introduction}
\label{sec:intro}
Collecting human demonstrations via teleoperation has become a popular paradigm for enabling robots to acquire task-specific skills~\cite{chi2023diffusionpolicy, lee2024behavior, zhao2023learning}. Although such demonstrations can effectively bootstrap imitation learning policies, their deployment often reveals critical limitations. Due to compounding errors during deployment and the stochasticity of environments, the learned policy often falls into out-of-distribution (OOD) states where the policy struggles to generalize suitable actions. While collecting additional data with careful randomization may increase the coverage of states in the demonstrations, the approach still suffers from low data efficiency in the absence of prior knowledge about the OOD states.

Several recent efforts have attempted to address this challenge through human-in-the-loop data augmentation, in which the robot runs automatically most of the time while human intervention is introduced when the policy fails in order to provide corrective demonstrations. These corrective behaviors are subsequently incorporated into the demonstration dataset, thereby directly guiding the robot on how to act in OOD states.

Sirius~\cite{liu2022robot}, for example, leverages a space mouse for human intervention and correction to enable mixed control between a learned policy and human teleoperation during deployment, while simultaneously collecting new data for online fine-tuning. However, the corrective capabilities of the space mouse are inherently constrained—it only allows uniform translational or rotational adjustments, making it unsuitable for more complex refinements. Robo-Copilot~\cite{wu2025robocopilot} instead introduces a dual-robot setup in which a "follower" robot copies joint positions of a teleoperated "leader" robot for human demonstrations while the leader mirrors the follower robot during policy execution. This design benefits from the shared workspace and kinematic equivalence of the two robots, enabling intuitive data collection during deployment. Yet, the requirement of identical or proportionally scaled kinematics across robots fundamentally limits its applicability and prevents broader generalization.

These approaches highlight the tension between dexterity and generality in human-in-the-loop robot learning. Although existing systems provide useful correction methods, they struggle to support fine-grained control while remaining broadly compatible with heterogeneous robot platforms at the same time.

To address these limitations, we propose Human-Robot Copilot, a cross embodiment framework designed to improve the entire pipeline of human demonstration collection and imitation learning. Our system ensures that the leader and follower robots share overlapping workspaces, enabling intuitive corrective teleoperation during deployment. At the same time, heterogeneity in arm design allows us to introduce scaling factors that facilitate fine-grained control, while providing full 6-DoF end-effector pose commands that are compatible with a wide range of industrial and research manipulators.

We demonstrate that fine-tuning with human data collected through this framework significantly improves policy performance on contact-rich, high-precision, and logically complex tasks. This highlights the potential of heterogeneous teleoperation systems to bridge the gap between efficient human data collection and robust robot learning in real-world deployment scenarios. The complete system, including hardware design, teleoperation software, and training pipeline,
is open-source. The leader device is built from commercially available components and 3D-printed
parts at a total cost of approximately \$1{,}000, enabling broad adoption and reproducibility.

In summary, our contributions are:
\begin{itemize}[leftmargin=*,nosep]
  \item An integrated, open-source human-in-the-loop framework that bridges
        heterogeneous teleoperation and deployment-time policy refinement at
        a hardware cost of approximately \$1{,}000.
  \item A bidirectional leader--follower control architecture with adjustable
        scaling that enables seamless switching between policy execution and
        fine-grained human intervention across diverse manipulators.
  \item Experimental validation on both simulated and real-world tasks showing
        that the proposed framework achieves higher task success rates with
        the same number of demonstrations while requiring substantially less
        data collection time.
\end{itemize}

\section{Related Work}
\label{sec:related}

\subsection{From offline imitation learning to human-in-the-loop.}  Traditional imitation learning paradigms~\cite{chi2023diffusionpolicy, yan2025maniflow, brohan2022rt, zitkovich2023rt, haldar2023teach} typically follow a three-stage pipeline: collecting a fixed set of human demonstrations, training a policy on this dataset, and subsequently deploying the learned policy. Once the demonstrations are provided, the human supervisor is excluded from the control loop, leaving the policy fully responsible for execution. In contrast, human-in-the-loop imitation learning~\cite{mandlekar2020human, liu2022robot, wu2025robocopilot, spencer55learning, kelly2019hg} maintains continuous human involvement during deployment, enabling supervision, intervention, supplementary demonstrations, and evaluation. This paradigm not only addresses data sparsity by allowing humans to provide additional demonstrations on failure cases~\cite{kelly2019hg, spencer55learning} but also enhances safety through real-time teleoperation override~\cite{liu2022robot}. Moreover, it improves data efficiency by focusing human input on the most challenging parts of the task where the current policy underperforms. As the policy improves, the human workload can gradually decrease. A key requirement for these systems is the ability to seamlessly switch between autonomous policy control and human teleoperation, which fundamentally depends on aligning the workspaces of the human and robot. However, existing teleoperation devices often fall short of this requirement~\cite{liu2022robot, luo2025precise, mandlekar2020human}.

\subsection{Teleoperation devices.}
Current human-in-the-loop teleoperation interfaces largely fall into two categories. The first category, exemplified by Smartphone, SpaceMouse and VR controllers~\cite{mandlekar2020human, liu2022robot, luo2024serl, zhang2018deep, cheng2024open, dass2024telemoma}, provides cross embodiment control. Here, the human input is added as a displacement or velocity increment to the current robot states. While these devices are general-purpose and capable of both coarse and fine motions, the delta control is unintuitive and ill-suited for tasks requiring both large-scale movements and fine-grained precision, since the controller’s displacement is relative and lacks a fixed spatial reference to the robot’s base, making it difficult for operators to perceive absolute position or scale. Moreover, demonstrations collected via these devices show higher variance and jerk, making them harder for the policy to learn. 

The second category relies on joint-space mirroring using paired, isomorphic robot arms~\cite{zhao2023learning, fu2024mobile, wu2024gello, wu2025robocopilot, liu2025factr, ze2025twist}. These devices enable intuitive one-to-one control by directly mapping joint angles, but their limitations are twofold: (1) the teleoperation scale is fixed, making it difficult to perform fine-tuning in precision-demanding subtasks, and (2) their homomorphic design restricts transferability, as a leader arm can only control a structurally identical follower. Adapting to a new robot therefore requires redesigning and rebuilding the leader hardware, significantly reducing flexibility.

\begin{figure*}[t]
  \centering
  \includegraphics[width=\textwidth]{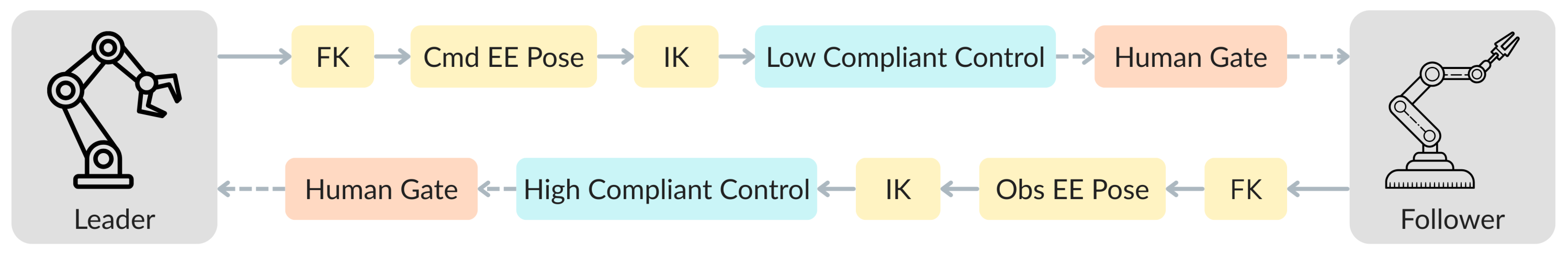} % 占满两栏宽
  \caption{Bidirectional control and observation communication. Forward and inverse kinematics (FK/IK) are continuously computed for both the leader and follower robots. Dashed lines denote control signals, of which only one is selected for synchronization between the two robots. The human teleoperator determines which control signal the two robots execute.}
  \label{fig:communication}
\end{figure*}

\subsection{Towards cross-embodiment copilots.}
To overcome these limitations, we propose a cross embodiment copilot framework that integrates hardware, control, and learning. By leveraging kinematic workspace alignment rather than strict joint homomorphism, humans can perform both large-scale and fine-grained control across diverse robot morphologies. Moreover, the framework allows humans to efficiently provide supplementary demonstrations and fine-tuning data during deployment, directly targeting the failure modes of the current policy. This design achieves the intuitiveness of joint-space mirroring with the generality lacking in existing approaches, while closing the loop between control and learning to enable data-efficient human-in-the-loop imitation learning.

\section{Method}
\label{sec:method}
Our framework consists of two main components: an interactive heterogeneous teleoperation system and a human-in-the-loop imitation learning pipeline. We first introduce the architecture and usage of the teleoperation system, which enables interactive data collection across heterogeneous embodiments at different control scales. This system is used to collect both the initial demonstrations and the fine-tuning data during deployment. We then describe how the collected fine-tuning data are integrated into the imitation learning process to continually improve the policy.

\subsection{Teleoperation System for Human-in-the-loop}
\label{sec:teleop}

To realize a cross-embodiment teleoperation system, a common approach is to leverage the inverse-kinematic (IK) solver to translate the desired end-effector pose derived from the leader device into the joint positions of the follower arm. While prior IK-based cross-embodiment devices such as ACE~\cite{yang2025ace} demonstrate the viability of decoupling leader and follower kinematics, they are designed as wearable exoskeletons for unidirectional offline data collection. In contrast, our system employs a compact, low-cost tabletop leader device with 3-DoF translational force feedback, and critically introduces bidirectional control that enables seamless mode switching between autonomous policy execution and human teleoperation---a capability essential for human-in-the-loop deployment but absent in existing IK-based designs~\cite{cheng2024open, yang2025ace}. Setting up cross-embodiment alignment requires only loading the follower robot's URDF and specifying the end-effector link, with no task-specific calibration beyond defining overlapping reachable workspaces; residual mapping deviations are corrected online through visual closed-loop feedback and bidirectional synchronization.

Our control logic operates in two modes. During the initial data collection, the leader arm reads the motor encoders and IMU signals. The joint readings are passed through forward kinematics to obtain the end-effector position, which is then combined with the IMU rotation to estimate the full end-effector pose. To maximize mechanical transparency and maintain low inertia for highly dexterous movements, the leader device focuses on 3-DoF translational force feedback, while the 3-DoF rotational tracking is purely achieved via IMU mapping without active
torque feedback, ensuring that orientation control remains lightweight and does not
over-constrain the leader's mechanical design. This pose is provided to the IK solver of the follower arm, and the resulting command joint positions are sent to the follower arm for execution. In parallel, the follower returns observed joint positions, which are converted via forward kinematics into Cartesian end-effector positions. During data collection, these states of the follower are only treated as observations.

\begin{figure}[t]  % 单栏
  \centering
  \includegraphics[width=0.48\textwidth]{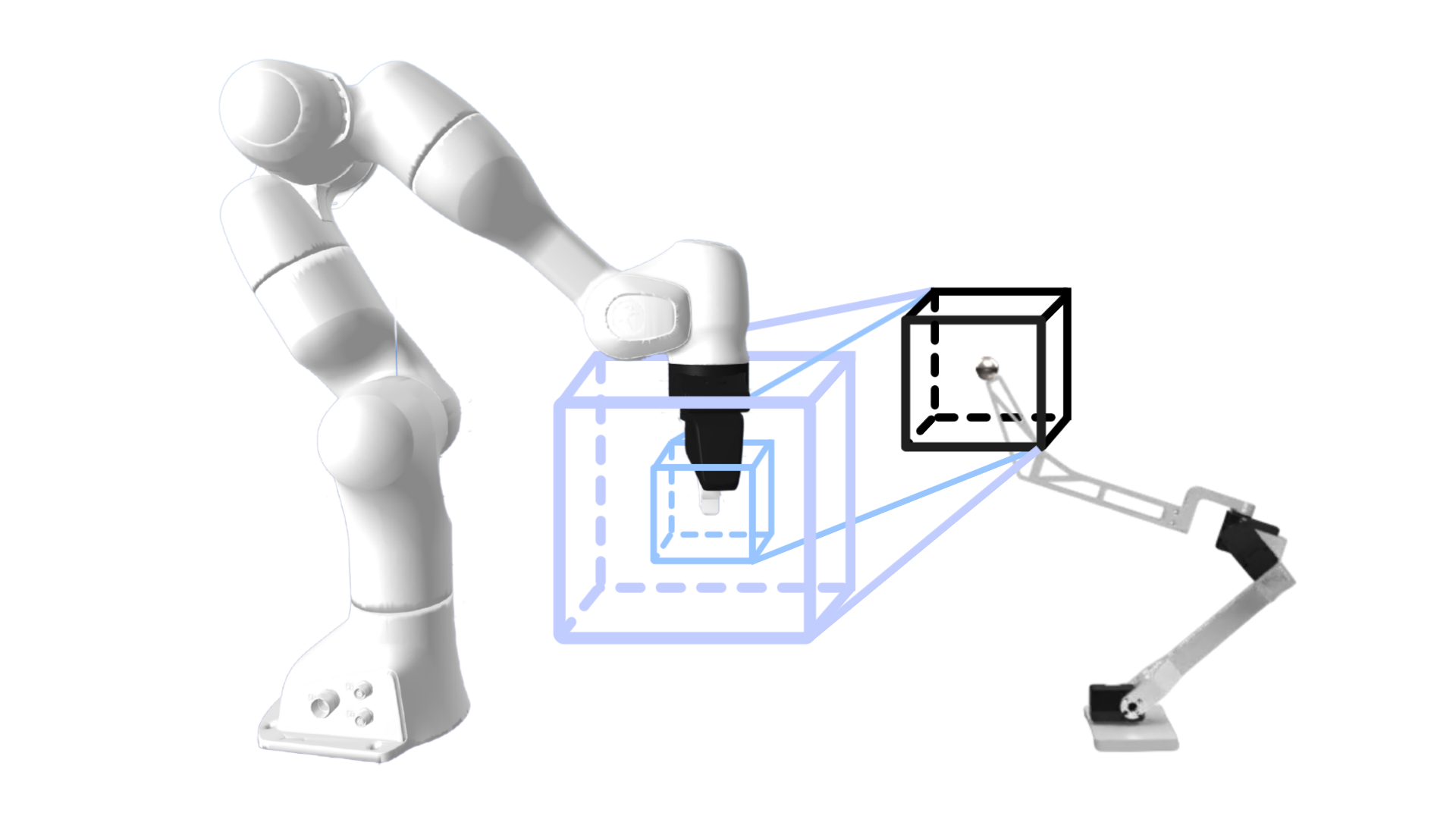} % 占半栏宽
  \caption{Illustration of task workspaces under different scaling factors. The black cube represents the workspace of the leader arm, while the two blue cubes correspond to task workspaces under different scaling factors. A larger task workspace facilitates rapid large-scale movements, whereas a smaller task workspace supports precise and accurate actions for high-precision tasks.}
  \label{fig:scaling}
\end{figure}

During human-in-the-loop fine-tuning, teleoperation and policy execution run as two parallel channels, although only one channel’s output is forwarded to the follower arm at any given time. The operator can switch between teleop and policy channels with a single command. Importantly, because of the bidirectional control logic, even when the follower listens to the policy channel, the leader arm continues to receive the follower’s joint positions. These joint positions are converted via forward kinematics into Cartesian end-effector pose, and then solved again through IK to update the joint positions of the leader arm, ensuring both arms remain aligned in the same workspace. This makes switching control intuitive for the human operator, as continuous synchronization
preserves proprioceptive awareness of the follower's state and ensures sufficient correction
range of motion. Meanwhile, both channels can be subscribed by the data collection program, which records synchronized command joint positions, observed joint positions, and images under a unified timestamp to construct data clips. Furthermore, when running the policy, switching into teleoperation mode allows us to compare the policy outputs against human demonstrations, thereby inspecting the abnormal output of the policy and identifying potential out-of-distribution (OOD) states.

To stabilize control, both leader and follower arms are equipped with gravity and friction compensation, allowing them to be driven with low-gain PD controllers. In teleoperation mode, the returned control commands use even lower PD gains to suppress oscillations and improve compliance. In policy mode, by contrast, friction compensation of the leader arm is disabled and its PD gains are increased, ensuring real-time synchronization via high-frequency wired connections between the two end-effectors. The leader device communicates with the control computer via USB, while the follower arm is                             
  controlled through a wired Ethernet connection using the libfranka FCI interface. Both the                              
  teleoperation control loop and inter-process communication via LCM run locally at high                                  
  frequency, ensuring negligible synchronization delay. The bidirectional communication pipeline is illustrated in Fig.\ref{fig:communication}.

\subsection{Control Scale Adjustment}

By leveraging the heterogeneous characteristics of the system, the teleoperation can be flexibly adjusted, enabling coarse control for large-scale movements and fine-grained control for precision-demanding tasks. The illustration of the task workspaces under different scaling factors is in Fig.\ref{fig:scaling}.

Since the orientation of the end-effector does not require scaling, we consider only the positional components. Let follower end-effector position be $\mathbf{x}_f \in \mathbb{R}^3$, and the leader end-effector position be $\mathbf{x}_l \in \mathbb{R}^3$, the mapping between the two positions is defined as
\begin{equation}
    \mathbf{x}_f = \alpha (\mathbf{x}_l - \mathbf{c}_l) + \mathbf{c}_t,
\end{equation}
where $\mathbf{c}_l$ denotes the center of the leader robot's workspace, $\mathbf{c}_t$ denotes the center of the task workspace, and $\alpha$ is the scaling factor.

For large-scale tasks, such as object transfer, we employ a larger scaling factor ($\alpha=2.0$) for the alignment between the two workspaces. Conversely, for precision-demanding tasks, such as object insertion, a smaller scaling factor ($\alpha=0.5$) is adopted to enable more accurate teleoperation. The scaling factor is an operator-level preference rather than a task-specific hyperparameter, and can be adjusted without retraining the policy. We use fixed scaling factors because continuously varying $\alpha$ would break the absolute pose mapping during bidirectional mode switching.

\subsection{Human-in-the-loop Imitation Learning}

\begin{figure}[t]
  \centering
  \includegraphics[width=0.45\textwidth]{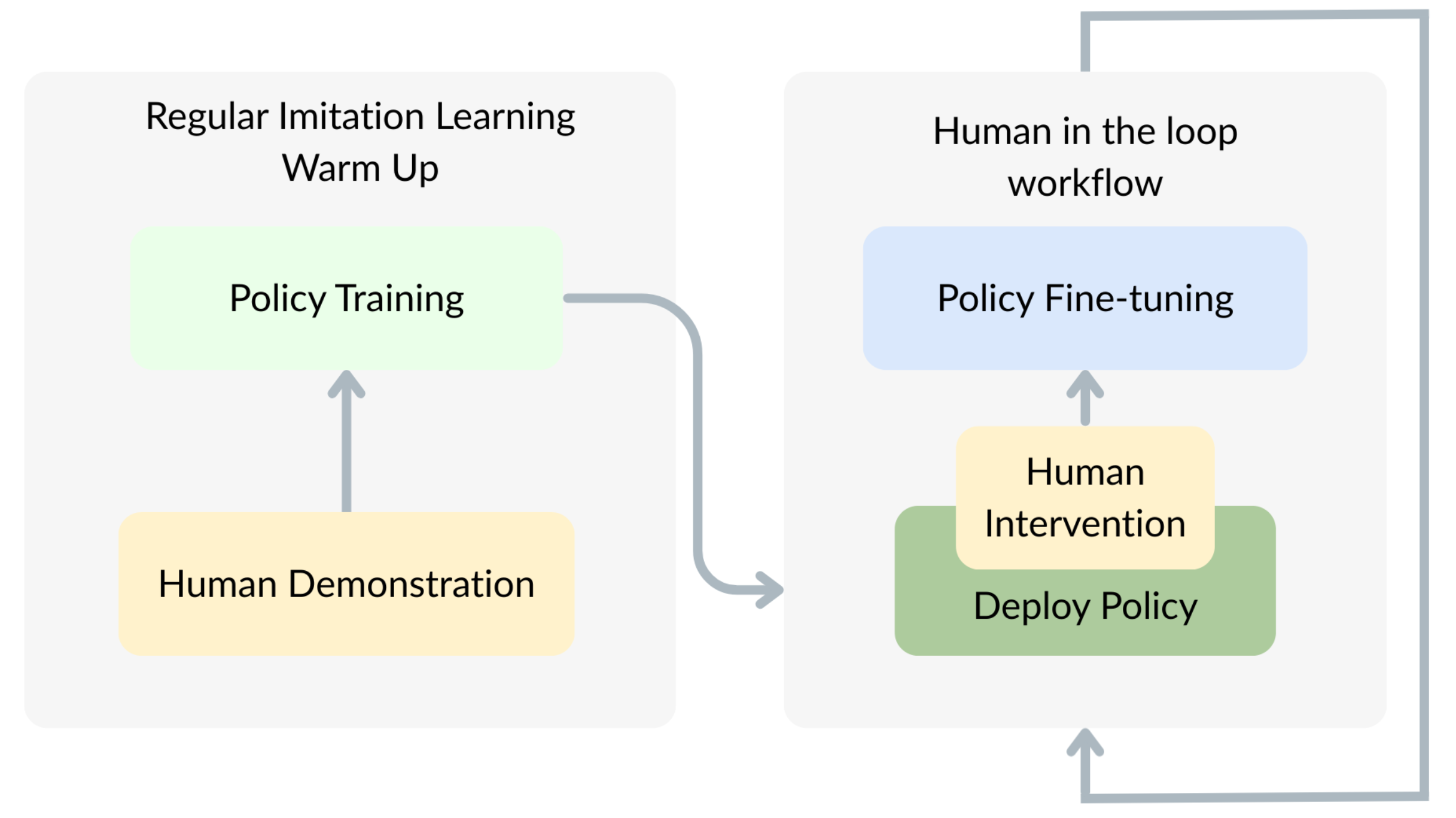} % 占满两栏宽
  \caption{Training and data augmentation workflow. The base policy is first initialized through regular imitation learning. It is then deployed to identify potential failure modes. During deployment, a human teleoperator intervenes when necessary, providing corrective actions. These corrective demonstrations are recorded and incorporated into the original dataset, which is subsequently used to fine-tune the policy.}
  \label{fig:training}
\end{figure}

After obtaining the base policy, we proceed to human-in-the-loop data collection. The training and data augmentation workflow is illustrated in Fig.\ref{fig:training}. During deployment, the robot executes the learned policy, but the human operator can take over control via teleoperator at any time. When intervening, the operator can also selectively record demonstrations. Unlike full trajectories, we only log the segments after human takeover, which we call data clips. These clips may correspond to different parts of a task rather than entire demonstrations, but they provide targeted supervision precisely where the policy struggles.

Compared to traditional offline training pipelines such as Action Chunking Transformer (ACT)~\cite{zhao2023learning}, human-in-the-loop imitation learning requires the policy to support fast iterations. Therefore, we utilize the ACT policy while reducing the network size to enable rapid finetuning after data augmentation. Moreover, we choose ResNet-18 as our vision backbone instead of other more powerful but larger choices, such as Dino-V2, used by other ACT-based policies~\cite{cheng2024open}. Note that the proposed data collection and fine-tuning pipeline is policy-agnostic; ACT is adopted here for its fast iteration capability, and can be replaced by other imitation learning policies that support incremental training.

For fine-tuning, we combine the collected data clips with the original demonstrations (to avoid catastrophic forgetting of previously learned skills). We summarize the fine-tuning procedure in Algorithm 1. With the smaller network, training the base policy from scratch takes about 40 minutes while finetuning takes less than 10 minutes to converge, enabling us to rapidly redeploy the updated policy. This fast retraining cycle supports repeated rounds of human-in-the-loop data collection and refinement, making the framework practical for real-world iterative improvement.

\begin{algorithm}[t]
\caption{Human-in-the-loop Imitation Learning with Clip-Based Batch Finetuning}
\label{alg:hil_clip_simple}
\begin{algorithmic}[1]
\Require Base demos $\mathcal{D}_0$, initial policy $\pi_0$, human expert $\pi^\ast$, trigger $K$, iterations $N$
\State \textbf{Base train:} $\pi_1 \leftarrow \text{BC}(\mathcal{D}_0)$
\For{$i=1$ to $N$}
  \State $\mathcal{C} \gets \emptyset$ \Comment{clip buffer}
  \State \textbf{Deploy} $\pi_i$
  \While{task not done}
    \State observe $s$
    \If{human intervenes}
      \State start clip $C \gets \emptyset$
      \While{human in control}
        \State execute $a^\ast \gets \pi^\ast(s)$; \, append $(s,a^\ast)$ to $C$; \, step env; \, update $s$
      \EndWhile
      \State $\mathcal{C} \gets \mathcal{C} \cup \{C\}$
    \Else
      \State execute $a \gets \pi_i(s)$; \, step env; \, update $s$
    \EndIf
    \If{$|\mathcal{C}| \ge K$}
      \State $\mathcal{D} \gets \mathcal{D}_0 \cup \bigcup_{C \in \mathcal{C}} C$
      \State $\pi_{i+1} \leftarrow \text{Finetune}(\pi_i,\mathcal{D})$; \, $\mathcal{C} \gets \emptyset$; \, \textbf{Redeploy} $\pi_{i+1}$
    \EndIf
  \EndWhile
\EndFor
\State \textbf{Output:} final policy
\end{algorithmic}
\end{algorithm}

\section{Experiments}
\label{sec:exp}

\begin{figure*}[t]
    \centering
    \includegraphics[width=\textwidth]{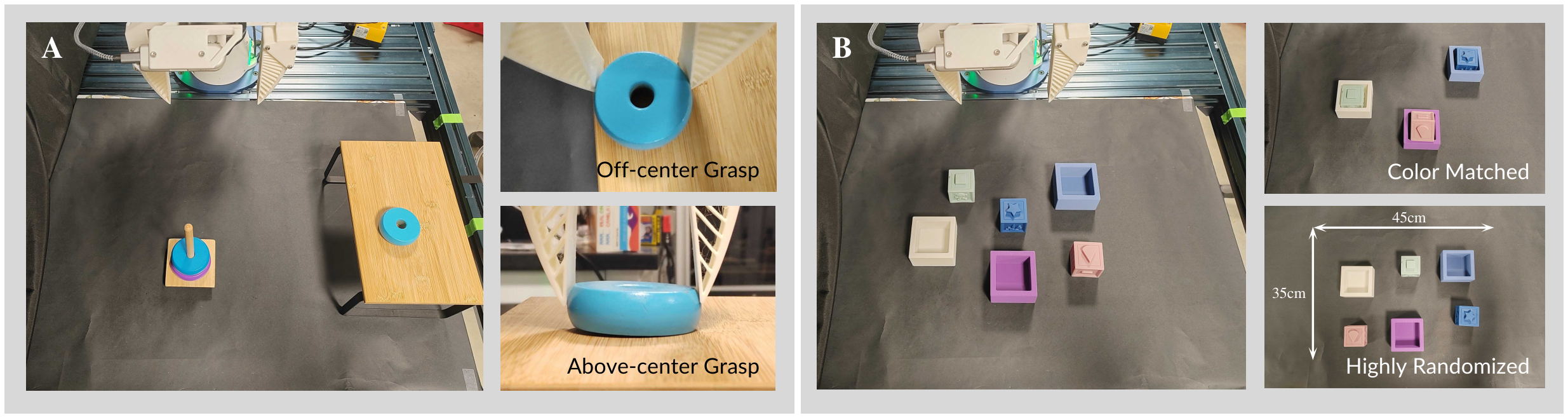}
    \caption{The real-world experiments and their key challenges. Fig. A illustrates the tower of hanoi insertion task. In addition to the narrow tolerance required for insertion, grasping the disk itself is challenging. From the top view (top-right), the gripper must align precisely with the disk’s center; otherwise, the disk slips out. From the side view (bottom-right), the gripper must also engage below the midpoint of the disk’s curved edge. Fig. B presents the cube sorting task, where cubes of different colors must be placed into their corresponding containers. The six objects are randomly distributed within a 45 cm × 35 cm workspace, creating a highly randomized environment that significantly increases the difficulty of learning correct actions.}
    \label{fig:real_exp}
\end{figure*}

The experimental setups are illustrated in Fig.~\ref{fig:tease}. We conducted a series of experiments to demonstrate the capabilities of the proposed framework to locate the states where the policy outputs abnormal actions and then collect targeted fine-grained demonstrations with different scaling factors. This therefore shows the data efficiency of the framework. 

All policies are initialized with the same set of human demonstrations as a warmup. While more demonstrations are then added to the dataset of the base policy to train it again from scratch, we augment the dataset with our proposed Human-Robot Copilot and ensure that the total number of trajectories matches that of the base policy for a fair comparison.

\subsection{Tasks}

\subsubsection{Simulation}
We conducted two simulation experiments on the standard Robomimic benchmark~\cite{robomimic2021} to validate the effectiveness of our proposed framework in a controlled reproducible environment. We choose PickPlace(Can) and NutAssembly(Square) tasks from the benchmark. 

For the can pick-and-place task, precise grasping is required. If the gripper fails to clamp the can at its center, the cylindrical structure of the can leads to uneven force distribution, causing the object to gradually slip during transportation. 

In the square nut assembly task, randomization of the nut’s orientation and position increases the diversity of possible states, thereby requiring a larger amount of demonstrations to ensure sufficient coverage. Moreover, successful assembly demands highly precise placement, further compounding the difficulty of the task. In order to generate relatively easy demonstrations for the policy to learn, we adopted a two-stage procedure: first orienting the nut correctly, followed by grasping. In detail, if the nut was already in the right direction, a grasping action was directly executed. Otherwise, the nut was rotated incrementally—by up to 90 degrees per step—until it reached the correct orientation, after which the grasping action was performed. With this method, the policy does not need to learn to generate right grasp actions for all orientations of the nut.

\subsubsection{Real World}
The two real-world tasks are selected to evaluate human-in-the-loop intervention under two distinct classes of failure modes. We designed a cube sorting task with a wide range of randomization to increase the likelihood of encountering out-of-distribution (OOD) states, as well as a tower of hanoi insertion task to evaluate the framework’s capability in executing precise, contact-rich actions. The experimental setup and key challenges are illustrated in Fig.\ref{fig:real_exp}.

For the cube sorting task, we randomly placed the three cubes with different colors and three corresponding square containers in a 45cm $\times$ 35cm area. The robot is required to pick those cubes and then place them into the right container. Since the six objects have a wide range of randomized positions, there is a large amount of data needed to cover all the possible states. 

For the tower of hanoi insertion task, the blue disk to be inserted is randomly placed on the table, while the tower itself is positioned along a line with a randomization range of 3 cm. Since the task is designed primarily for evaluating the ability of performing precise actions, we did not choose a wide range for randomization. The physically measured diameter of the tower’s pole is 13.6 mm, whereas the diameter of the disk’s central hole is 15.6 mm, resulting in a tolerance margin of only 2 mm for successful insertion. Under this condition, the 3 cm randomization is sufficient to ensure that the learned policy is genuinely reasoning about how to place the disk, rather than merely replaying the placement trajectories observed in the human demonstrations.

\begin{table*}[t]
\centering
\resizebox{0.78\textwidth}{!}{
\begin{tabular}{lccccccc}
\toprule
\textbf{Task} & \textbf{Num of Traj} 
& \multicolumn{3}{c}{\textbf{Base Policy}} 
& \multicolumn{3}{c}{\textbf{Proposed}} \\
\cmidrule(r){3-5} \cmidrule(r){6-8}
 &  & Stage 1 & Stage 2 & Total 
 & Stage 1 & Stage 2 & Total \\
\midrule

\multirow{2}{*}{Can Pick \& Place} 
 & 20 (warmup) & 60 & 86 & 52 & -- & -- & -- \\
 & 40 & 86 & 74 & 64 & 84 & 80 & \textbf{72} \\
\cmidrule(r){1-8}

\multirow{2}{*}{Nut Assembly}       
 & 20 (warmup) & 30 & 84 & 26 & -- & -- & -- \\
 & 40 & 46 & 94 & 46 & 56 & 100 & \textbf{56} \\

\bottomrule
\end{tabular}
}
\caption{Simulation results. Each task is divided into two stages during evaluation. Each task is evaluated over 50 rollouts. We report success rate (\%) for each stage and overall performance.}
\label{tab:sim-success-rates}
\end{table*}

\begin{table*}[t]
    \centering
    \resizebox{\textwidth}{!}{
\begin{tabular}{lcccccccccccccc}
\toprule
\textbf{Task} & \textbf{Num of Traj} 
& \multicolumn{3}{c}{\textbf{Base Policy}} 
& \multicolumn{3}{c}{\textbf{Keyboard Policy}}
& \multicolumn{3}{c}{\textbf{VR Policy}}
& \multicolumn{3}{c}{\textbf{Proposed}} \\
\cmidrule(r){3-5} 
\cmidrule(r){6-8}
\cmidrule(r){9-11}
\cmidrule(r){12-14}
 &  
 & S1 & S2 & Total 
 & S1 & S2 & Total
 & S1 & S2 & Total
 & S1 & S2 & Total \\
\midrule

\multirow{2}{*}{Cube Sorting}              
 & 25 (warmup) & 54 & 46 & 24 & -- & -- & -- & -- & -- & -- & -- & -- & -- \\
 & 50 & 78 & 56 & 40 & 74 & 62 & 48 & 74 & 86 & 62 & 78 & 86 & \textbf{64} \\
\cmidrule(r){1-14}

\multirow{2}{*}{Tower of Hanoi Insertion}  
 & 25 (warmup) & 70 & 84 & 54 & -- & -- & -- & -- & -- & -- & -- & -- & -- \\
 & 50 & 68 & 90 & 60 & 72 & 88 & 66 & 90 & 78 & 74 & 90 & 92 & \textbf{86} \\

\bottomrule
\end{tabular}
    }
    \caption{Real-world results with additional teleoperation baselines (Keyboard and VR). Each task is evaluated over 50 rollouts. We report success rate (\%) for each stage and overall.}
    \label{tab:real-success-rates}
\end{table*}

\subsection{Experiment Results}

For each real-world task, we deploy the same base policy for the same number of rollouts
and use three different human-in-the-loop interfaces (Keyboard, VR controller,
and the proposed Copilot system) to perform corrective interventions. Keyboard serves as a discrete velocity-control interface representing the delta-control paradigm used by SpaceMouse-based systems~\cite{liu2022robot}. The corrective data collected
by each interface is then used to fine-tune the base policy independently. Results are reported in Table~\ref{tab:sim-success-rates}, Table~\ref{tab:real-success-rates}, and Table~\ref{tab:data-collection}. To better illustrate the advantages of the proposed framework, we divide each task into two stages. The first stage ends when the manipulator successfully grasps the target object (i.e. can, nut, cube, disk), while the second stage ends upon completion of the full task. Notably, even if the manipulator fails in the first stage, we intervene to reposition it at the beginning of the second stage and still evaluate its success rate of the second stage.

The results indicate that the proposed human–robot copilot consistently outperforms the base policy across most tasks. As shown in Table~\ref{tab:real-success-rates}, the proposed system consistently outperforms VR and Keyboard baselines. This is because VR controllers lack physical grounding, causing operator fatigue and loss of proprioceptive reference in millimeter-level tasks, while keyboards lack proportional input and spatial grounding, making fine corrections unintuitive. Furthermore, Table~\ref{tab:data-collection} demonstrates that the time required to collect corrective data is substantially lower than that needed to acquire full trajectories, highlighting the data efficiency of our framework. Among the human-in-the-loop interfaces, the proposed system achieves the lowest collection
time on both real-world tasks, indicating that the bidirectional leader--follower setup
enables more efficient corrective interventions than keyboard or VR-based alternatives.

Notably, the performance gap between the proposed system and VR/Keyboard baselines is more
pronounced in the precision-demanding Tower of Hanoi task than in cube sorting, suggesting
that the advantage of bidirectional leader--follower teleoperation is most significant when
fine-grained corrective interventions are required.

In the following two sections, we analyze the factors contributing to the superior data efficiency of the proposed framework. We attribute this efficiency to two key aspects: the ability to accurately identify failure conditions, and the capability to perform fine-grained, concise corrective interventions. Intuitively, the spatial distribution of trajectory data is non-uniform across time---certain segments exhibit high positional variance while others remain tightly clustered---and clip-based collection concentrates supervision on the high-variance, failure-prone regions rather than diluting it across the entire trajectory.

\begin{table}[t]
    \centering
    \footnotesize
    \setlength{\tabcolsep}{3pt}
    \begin{tabular}{lccccc}
    \toprule
    Task & Added & Base & Key. & VR & Prop. \\
    \midrule
    Can Pick-and-Place & 20 & 501.7 & -- & -- & 189.8 \\
    Nut Assembly & 20 & 713.7 & -- & -- & 313.3 \\
    Cube Sorting & 25 & 1233.0 & 334.7 & 266.9 & 252.9 \\
    Tower of Hanoi Insertion & 25 & 474.2 & 260.7 & 333.3 & 227.4 \\
    \bottomrule
    \end{tabular}
    \caption{Total data collection time (s) using different human-in-the-loop interfaces.}
    \label{tab:data-collection}
\end{table}

\subsection{Locating Failure Conditions}

Through experiments, we identified failure conditions of the base policy and collected targeted
  demonstrations. Several failure cases were unexpected, suggesting that blind data augmentation without
  knowledge of such conditions may be insufficient.

For the can pick-and-place task in simulation, the primary challenge lies in accurately clamping the can at its center. Failures observed during Stage 2 (placing the can in the target location) are primarily a consequence of imprecise grasps in Stage 1, which cause the can to slip from the gripper during transfer.

For the cube sorting task in real world experiments, as we illustrated in Fig.\ref{fig:real_exp}, since there is a wide range of randomization of both the cubes and the containers, it is also obvious that the robot will struggle to grasp a cube or put it in the container that is located in a seldom visited place in human demonstrations. During deployment, the policy’s difficulties effectively indicate the out-of-distribution (OOD) states, allowing the collected augmentation data to be targeted specifically at these challenging states rather than covering the entire state distribution.

While the failure modes of the two above tasks are quite easy to expect, the policy failed unexpectedly in the other two tasks. Both of the other two tasks require the policy to generate precise actions for assembly or insertion, however, from Table~\ref{tab:sim-success-rates} and Table~\ref{tab:real-success-rates} we can see that it is not difficult for the policy to successfully insert the nut or disk, but it is hard to pick them.

For the nut assembly task, although the rotation procedure simplifies the demonstrations and facilitates policy learning, it can also introduce unintended shifts of the nut, potentially leading to the OOD states for the policy. For the tower of hanoi insertion task, the near-cylindrical geometry of the disk requires the same precise actions as the can pick-and-place task in the simulation. Moreover, the curved side edges of the disk require the gripper to grasp below its widest diameter, or the disk is prone to slipping during transfer.

\begin{figure}[t]
    \centering
    \includegraphics[width=0.45\textwidth]{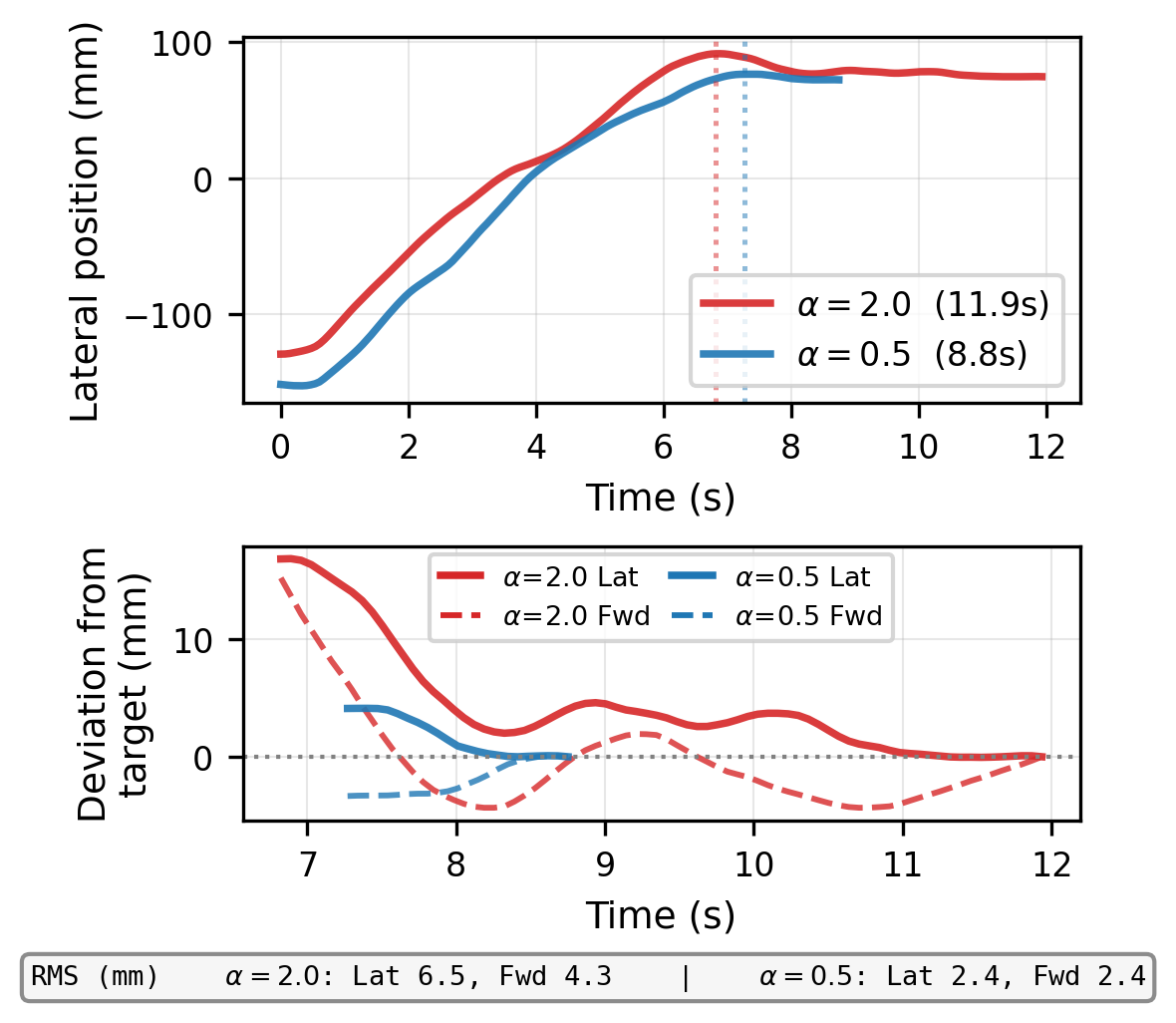}
    \caption{End-effector trajectories under different scaling factors ($\alpha=2.0$ vs $\alpha=0.5$) during Tower
  of Hanoi insertion. Top: lateral position over time showing the full transport-to-alignment trajectory.
   Bottom: lateral and forward deviations from the target position during the alignment phase (starting
  from the first velocity zero-crossing). The alignment-phase RMS deviations are reported below.}
    \label{fig:scale_result}
\end{figure}

\subsection{Fine-grained Correction}

When abnormal actions occur—such as attempting to grasp an object without proper alignment, failing to initiate the next action, or overshooting the intended target location—we intervene to provide corrective demonstrations that guide the policy.

For tasks requiring high precision, we observed that even human teleoperators face considerable challenges. For instance, in the Tower of Hanoi insertion task, the human teleoperator achieved only about a 40\% success rate under the default scaling factor. While failed demonstrations can be discarded during the collection of training data for the base policy, failure cases encountered during base policy deployment cannot be reliably reproduced, hence the correction demonstration is expected to succeed in a single attempt. This highlights the need to refine the policy using more precise teleoperation, underscoring the importance of adopting a smaller scaling factor.

We illustrate the end-effector trajectories collected under different scaling factors in
  Fig.\ref{fig:scale_result}. With $\alpha=2.0$, the operator reaches the target region faster but
  overshoots and oscillates during alignment, resulting in an RMS deviation of 6.5\,mm (lateral) and
  4.3\,mm (forward). In contrast, $\alpha=0.5$ produces a smooth, monotonic approach with an RMS deviation
   of only 2.4\,mm in both directions, confirming that a smaller scaling factor enables more precise
  alignment.

With $\alpha=0.5$, corrective demonstrations achieved a higher success rate and fewer attempts,
  reducing collection time and yielding more concise, effective training data.

\section{Conclusion}

We proposed the Human-Robot Copilot framework, compatible with diverse manipulators, leveraging a      
  scaling factor for dexterous teleoperation in high-precision tasks. Experimental results demonstrate 
  that human-in-the-loop data augmentation effectively identifies failure conditions and                 
  out-of-distribution (OOD) states, guiding the teleoperator to collect targeted demonstrations. The   
  scaling factor further enhances correction success in precision-demanding scenarios by enabling
  concise, easily imitable actions. Together, these enable data-efficient learning, requiring the same
  number of demonstrations—and less collection time—to outperform baselines trained with traditional
  methods.

Similar to how demonstration quality affects policy performance in traditional data collection pipelines, 
a limitation of the proposed framework—shared by HG-DAgger-style approaches—lies in the strategy for intervention 
and correction. Given the constraints of policy fine-tuning and ACT’s inability to handle multimodal demonstrations, 
correction actions should focus on refining the existing policy (e.g., more precise grasps or placements), rather than 
introducing entirely new execution modes. Training auxiliary policies on correction data, or adopting base policies 
capable of handling multimodality, may help address this limitation.

Finally, while fixed scaling factors are currently used to ensure control stability and data stationarity, 
exploring adaptive scaling mechanisms—such as adjusting the scale based on task phases or force feedback—remains 
a promising direction for future work, provided that workspace drift and mapping offsets can be compensated.

\bibliographystyle{IEEEtran}
\bibliography{main}

\end{document}